\def\year{2022}\relax
\documentclass[letterpaper]{article} % DO NOT CHANGE THIS
\usepackage{aaai22}  % DO NOT CHANGE THIS
\usepackage{times}  % DO NOT CHANGE THIS
\usepackage{helvet}  % DO NOT CHANGE THIS
\usepackage{courier}  % DO NOT CHANGE THIS
\usepackage[hyphens]{url}  % DO NOT CHANGE THIS
\usepackage{graphicx} % DO NOT CHANGE THIS
\usepackage{natbib}  % DO NOT CHANGE THIS AND DO NOT ADD ANY OPTIONS TO IT
\usepackage{caption} % DO NOT CHANGE THIS AND DO NOT ADD ANY OPTIONS TO IT
\DeclareCaptionStyle{ruled}{labelfont=normalfont,labelsep=colon,strut=off} % DO NOT CHANGE THIS
\usepackage{algorithm}
\usepackage{algorithmic}

\usepackage{newfloat}
\usepackage{listings}
\floatstyle{ruled}
\newfloat{listing}{tb}{lst}{}
\floatname{listing}{Listing}
\usepackage{multirow}
\usepackage{amssymb}

\title{VPFNet: Voxel-Pixel Fusion Network for Multi-class 3D Object Detection}
\author{
    Chia-Hung Wang,
    Hsueh-Wei Chen,
    Li-Chen Fu
}
\affiliations{
    Department of Computer Science and Information Engineering, National Taiwan University\\
}

\usepackage{bibentry}
\begin{document}

\maketitle

\begin{abstract}
\noindent Many LiDAR-based methods for detecting large objects, single-class object detection, or under easy situations were claimed to perform quite well. However, their performances of detecting small objects or under hard situations did not surpass those of the fusion-based ones due to failure to leverage the image semantics. In order to elevate the detection performance in a complicated environment, this paper proposes a deep learning (DL)-embedded fusion-based multi-class 3D object detection network which admits both LiDAR and camera sensor data streams, named Voxel-Pixel Fusion Network (VPFNet). Inside this network, a key novel component is called Voxel-Pixel Fusion (VPF) layer, which takes advantage of the geometric relation of a voxel-pixel pair and fuses the voxel features and the pixel features with proper mechanisms. Moreover, several parameters are particularly designed to guide and enhance the fusion effect after considering the characteristics of a voxel-pixel pair. Finally, the proposed method is evaluated on the KITTI benchmark for multi-class 3D object detection task under multilevel difficulty, and is shown to outperform all state-of-the-art methods in mean average precision (mAP). It is also noteworthy that our approach here ranks the first on the KITTI leaderboard for the challenging pedestrian class.
\end{abstract}

\section{Introduction}
3D Object detection plays a crucial role in autonomous driving, robotics, and augmented reality. Light detection and rangings (LiDARs) and RGB cameras are primary sensors to capture on-road semantic information.

\begin{figure}[t]
\centering
\includegraphics[width=0.9\columnwidth]{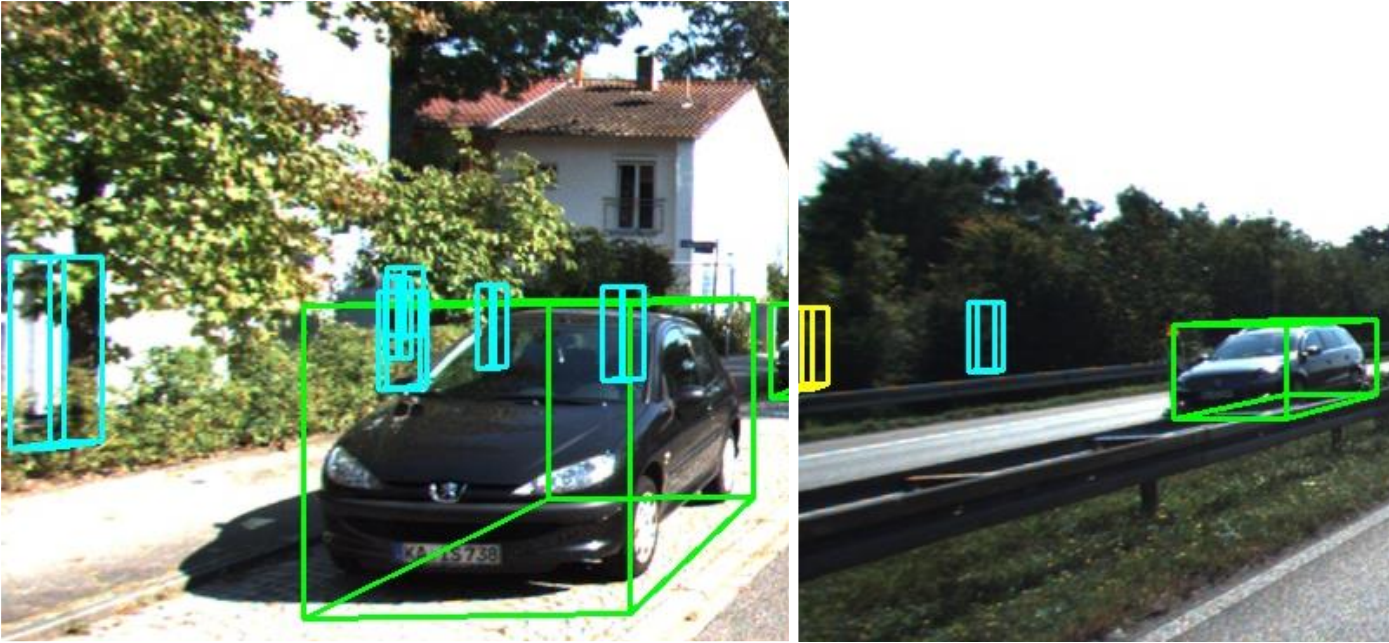} % Reduce the figure size so that it is slightly narrower than the column. Don't use precise values for figure width.This setup will avoid overfull boxes.
\caption{Challenges for LiDAR-based multi-class 3D detection. We observe the detection of PV-RCNN using their official code and pre-trained model. The green box denotes the correctly detected car. The blue ones and the yellow ones are false positives for pedestrians and cyclists, respectively.}
\label{fig:pvrcnnfp}
\end{figure}

Many LiDAR-based works \cite{pvrcnn, tanet} are developed in recent years because LiDARs capture inherent 3D information from point clouds, which appears to be promising at 3D object detection. However, the captured point clouds are sparse, discrete, and unordered, and they are sparser as the object distance becomes farther. Therefore, those works are limited if objects being far away need to be detected. Conceptually, without the help from visual semantic information, they cannot discriminate objects discriminated objects on small sizes with different categories or mixed with the background. As shown in Fig. \ref{fig:pvrcnnfp}, PV-RCNN \cite{pvrcnn} can correctly detect a car, but to detect small objects like pedestrians correctly remains challenging.

On the contrary, some camera-based works \cite{monofenet, dsgn} are actually deployed and put to use. Although cameras can capture high-resolution 2D images and sense objects’ colors and textures, they cannot acquire the depth information. Besides, they are passive devices and are susceptible to ambient light.

\begin{figure*}[t]
\centering
\includegraphics[width=0.95\textwidth]{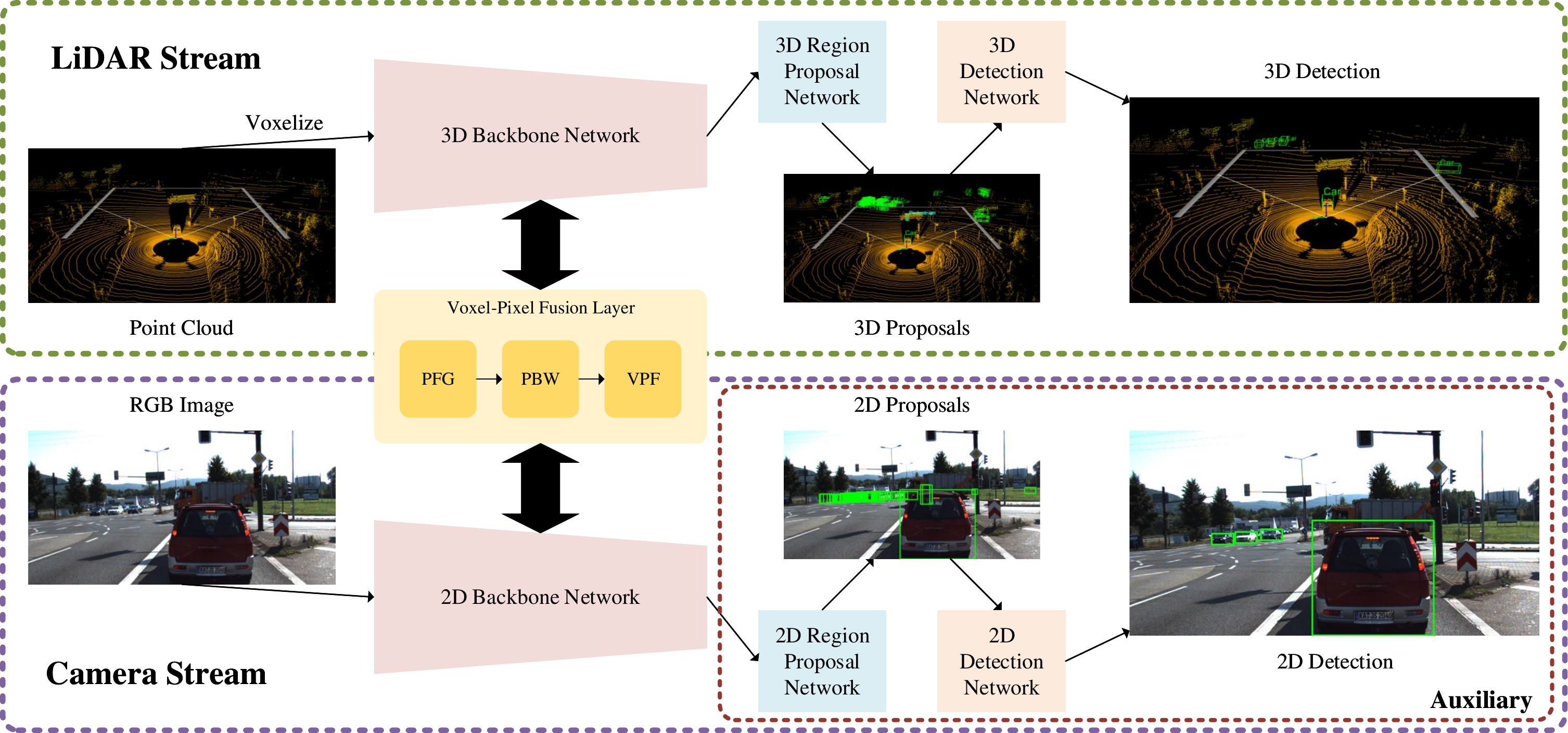} % Reduce the figure size so that it is slightly narrower than the column.
\caption{The overview of VPFNet. First, the LiDAR stream voxelizes the point cloud and utilizes a 3D backbone network to extract features. At the same time, the camera stream utilizes a 2D backbone network to extract features. Then, their feature maps are fed into the voxel-pixel fusion layer to fuse. Finally, the fused features are passed to the region proposal network and detection network to obtain the detection. Besides, the 2D detection task in the camera stream is auxiliary for training, which can be removed during inference time.}
\label{fig:VPFNet}
\end{figure*}

Since each sensor has its strengths and weaknesses, several fusion-based works \cite{mmf, pircnn, avod, mv3d} try to fuse the features from multiple sensors. However, there are different LiDAR point cloud representations since the point clouds are unordered and discrete. \cite{pircnn, farawayfrustum, fconvnet, fpointnet} directly fuse point-based features with the image, but the structure of a LiDAR point cloud is irregular, and hence cannot be directly admitted into a convolutional neural network. Therefore, several works \cite{mmf, avod, mv3d} have endeavored to cope with the challenge by using grid-based methods to transform the data format first so that the convolutional operations can be applied. But, it is noteworthy that LiDAR bird’s eye view (BEV) representation is another commonly adopted one used for fusion with features from the image. However, extracting features from BEV representation leads to serious loss of height information. Therefore, in this paper we try to fuse image pixel’s features with 3D voxel's features, rather than with the features extracted from the LiDAR BEV representation, since the voxel-wise representation can preserve most of the 3D information and is compatible with the convolutional operations.

This paper proposes a novel deep learning (DL)-embedded fusion-based 3D object detection network, called Voxel-Pixel Fusion Network (VPFNet) to cope with the aforementioned problems. The LiDAR point cloud representation is voxel-wise, which not necessarily preserves 3D space information but also admits convolutional operation. Our VPFNet takes LiDAR's point cloud and camera's RGB image as inputs, using Voxel-Pixel Fusion (VPF) layer to fuse two feature maps from respective sensors at feature-level. Finally, the voxel’s and pixel’s features in LiDAR and camera streams are fused bidirectionally. The experimental results on the KITTI \cite{kitti} dataset demonstrate the effectiveness of our approach. Besides, our approach outperforms the state-of-the-art works on detecting 3D pedestrian objects.

Contributions in this paper are summarized as follows:
\begin{itemize}
\item A novel two-stream 3D object detection network is introduced, called VPFNet, which takes LiDAR's point clouds and camera's RGB images as inputs and outputs 3D bounding boxes that enclose the detected objects.

\item A Voxel-Pixel Fusion layer is proposed to fuse voxel features and pixel features which takes both feature maps as inputs and outputs the fused feature maps. Furthermore, the geometric relation of a voxel-pixel pair is explicitly characterized. Moreover, Parameter Feature Generating (PFG), Parameter-Based Weighting (PBW), and Voxel-Pixel Fusion (VPF) modules are presented inside the VPF layer.

\item Several parameters are designed for the PFG module after considering voxel-pixel pair's characteristics, and the latent parameter features are generated for further use.

\item We conduct experiments on the KITTI dataset to verify the effectiveness of our proposed layer and modules. Lastly, our VPFNet has been shown to outperform the state-of-the-art methods.
\end{itemize}

\section{Related Work}
\subsection{Camera-Based 3D Object Detector}
For monocular-based methods, only one camera is used to predict 3D objects. MonoFENet \cite{monofenet} first estimates disparity from the input image and then converts the estimated disparity feature map into a virtual point cloud. Finally, the fused features of the input image and the virtual point cloud are applied to detect objects. 
As for stereo-based approaches, at least two or even more cameras serve to detect 3D objects. Thus, the spatial relationship between two cameras can provide depth information. DSGN \cite{dsgn} first uses a shared encoder to encode features from both stereo images, followed by construction of 3D geometric volume and plane-sweep volume. Lastly, DSGN predicts objects on BEV from 3D geometric volume.

\begin{figure*}[t]
\centering
\includegraphics[width=\textwidth]{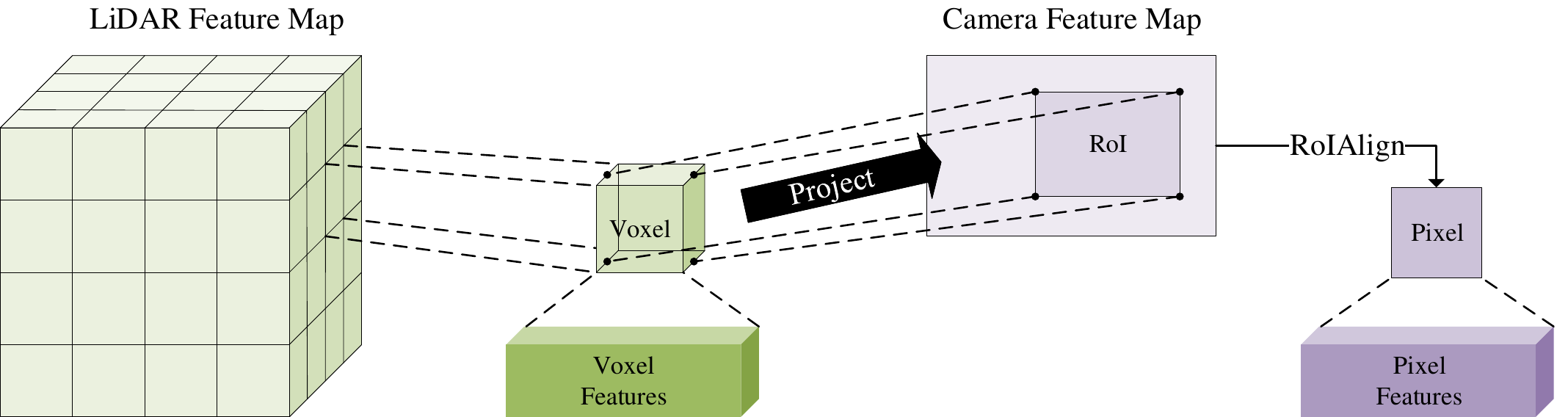} % Reduce the figure size so that it is slightly narrower than the column.
\caption{The process of finding voxel-pixel pair. First, a  non-empty voxel is picked up, and its voxel features are preserved. Second, four boundary points are chosen from each voxel. Third, these four boundary points will be projected onto the camera feature map, characterized in an RoI. Fourth, 1×1 RoIAlign is performed to get a pixel feature vector.}
\label{fig:VoxelPixelPair}
\end{figure*}

\subsection{LiDAR-Based 3D Object Detector}
The captured point clouds can be used to compute the relative distances to the ego-vehicle, and thus some researchers go ahead to use point cloud information to detect 3D objects and achieve commendable performance. However, since LiDARs collect unordered and discrete points, the raw representation cannot directly serve as the input to convolutional layers. Therefore, related methods can be divided into grid-based and point-based methods, respectively, according to different LiDAR representations.
As for PointNet \cite{pointnet}, which directly consumes raw points and considers the permutation invariance to conduct 3D object classification, part segmentation, and scene segmentation in terms of point-based methods.

\subsection{Fusion-Based 3D Object Detector}
According to our observation, most LiDAR-based methods surpass fusion-based methods because the significant number of objects are cars, whose sizes are often larger than cyclists and pedestrians. Several comparisons on smaller objects show that fusion-based methods do not perform worse than LiDAR-based methods. Another reason is that fusion-based methods usually require more computing resources, which sets a limit on the complexity of a fusion-based method. However, theoretically speaking, if either LiDAR-based or camera-based methods can achieve the highly promising performance, then combining these methods is supposed to manifest similar or even better results than single-sensor methods. Numerous approaches use different techniques to accomplish the fusion task. MV3D \cite{mv3d} takes LiDAR BEV representation, LiDAR front view representation, and the camera’s RGB image as inputs. It first generates 3D proposals on the BEV feature map, then projects them onto the other two feature maps, and finally fuses region-based features to generate the prediction. F-PointNet \cite{fpointnet} uses a mature 2D object detector to generate 2D proposals, which are then lifted into 3D space ones with a frustum. Lastly, the raw points inside the frustum are processed with PointNet to find the proper 3D objects. UberATG-MMF \cite{mmf} learns multiple interrelated tasks to make them complement one another mutually. The features learned from one task and one sensor can fused to benefit those learned from others. Despite various works have been proposed to tackle the sensor fusion problem such as those mentioned above, to the best of our knowledge, we are among the few groups of researchers who try to fuse voxel’s and pixel’s features from LiDAR and camera sensors, respectively, rather than to adopt the commonly used LiDAR BEV representation for serving the fusion purpose.

\section{VPFNet}
This section introduces our proposed novel 3D object detection network, named Voxel-Pixel Fusion Network (VPFNet), which consists of a voxel-pixel fusion layer (VPF layer) used to fuse features from multiple sensors. As shown in Fig. \ref{fig:VPFNet}, the VPFNet is composed of a LiDAR stream, a camera stream, and our proposed VPF layer.

Our proposed voxel-pixel fusion layer includes Parameter Feature Generating (PFG), Parameter-Based Weighting (PBW), and Voxel-Pixel Fusion (VPF) modules and gradually fuses voxel features and pixel features.

\subsection{Voxel-Pixel Fusion Layer}
The voxel-pixel fusion layer is a bridge to connect and enrich feature maps between LiDAR and camera streams. The inputs to the voxel-pixel fusion layer are the LiDAR feature map and camera feature map, which are the original feature maps originated from respective sensors. After fusion, the original feature maps will be replaced by the fused features maps, which are formed by summing the original feature maps and the fused features from the VPF layer. Since there are several down-sampling layers in backbone networks, the VPF layer can be created between one or more pairs of LiDAR feature map and camera feature map pair, but more VPF layers imply the requirement of longer computation time and may result in different final performance. In this work, we have done some ablation study to decide the optimal choice of layer number and setting, which will be revealed in the experiment section.

From the LiDAR feature map, we will find the corresponding pixel on the camera feature map, which can form a voxel-pixel pair. 
Because voxels are always sparser and pixels are always dense, an appropriate way to find a voxel-pixel pair is to start from each voxel in the LiDAR feature map. Using calibration to project a voxel onto the camera stream feature map, we can establish a pixel feature vector. The process of creating a voxel-pixel pair is illustrated in Fig. \ref{fig:VoxelPixelPair}. Because the area of the RoI on the camera feature map is not fixed and aligned, we utilize 1×1 RoIAlign \cite{maskrcnn} to transform the RoI into a pixel. Finally, each voxel-pixel pair can then be fed into the following modules in the VPF layer.

\subsubsection{Parameter Feature Generating Module.}
Since each sensor has different characteristics and the features for a voxel-pixel pair have some relation, it is better to fuse their features in terms of some parameters. Thus, we design a parameter feature generating module to generate parameter features to represent the deterministic characteristics of each voxel-pixel pair.

\subsubsection{Density parameter}
We define a density parameter $p_d$, which takes the number of raw points in each voxel into consideration. Moreover, the collected LiDAR points from a faraway place are often sparser than those from a nearer place. Thus, the impact of each feature may vary depending on its distance, far or near.

\subsubsection{Occlusion parameter}
We define an occlusion parameter $p_o$, which considers the number of occluding voxels. In other words, if a voxel occludes another one, the farther one will fall into an RoI, which is inside the image region onto which the nearer voxels are projected. Hence, pixel features may not provide enough or even correct information to the corresponding farther voxel features. As for how we calculate it, we first project all the voxels onto the image, and then check whether the projected RoI of each voxel is inside another from the other voxel and count the quantity.

\subsubsection{Area parameter}
Area parameter $p_a$ is the area of the projected RoI from a voxel, representing how many pixels a voxel corresponds to.

\subsubsection{Constrast parameter}
We define a contrast parameter $p_c$, which considers the contrast of the projected RoI in the raw image. That is to say, if the contrast of the RoI in the raw image is significant, the pixel features will describe diverse information as compared with those belonging to an inconspicuous RoI. The metric of contrast we adopt is Michelson contrast. The RGB image is first transformed into grayscale, and then the contrast of each RoI is calculated by $p_c=(I_{\rm max}-I_{\rm min})/(I_{\rm max}+I_{\rm min})$, where $I_{\rm max}$ and $I_{\rm min}$ stand for the maximum intensity and minimum intensity, respectively, in that RoI.

\subsubsection{Parameter feature generating}
\begin{figure}[t]
\centering
\includegraphics[width=0.7\columnwidth]{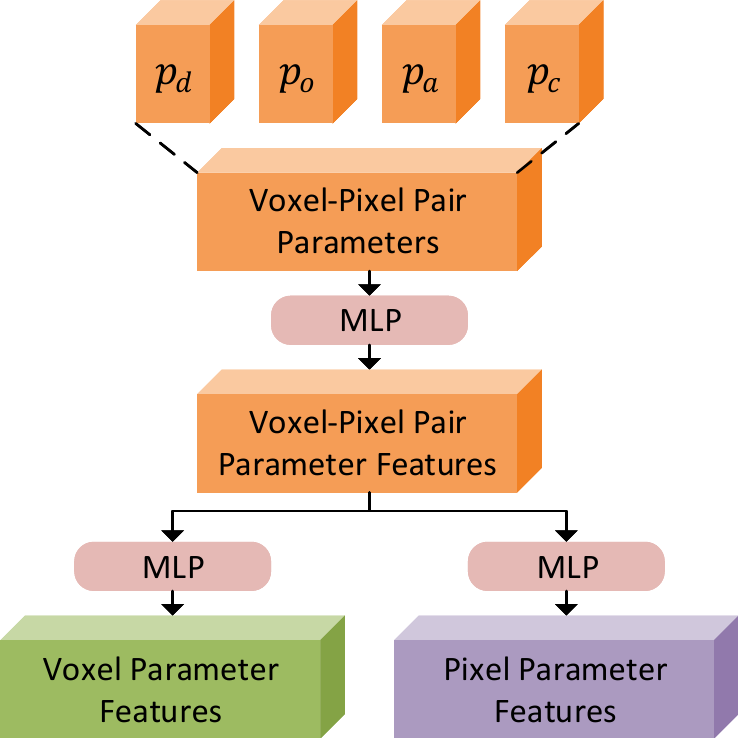} % Reduce the figure size so that it is slightly narrower than the column. Don't use precise values for figure width.This setup will avoid overfull boxes.
\caption{The structure of the Parameter Feature Generating module. First, the voxel-pixel pair parameters are transformed into pair parameter features using an MLP. Again, the features are transformed into voxel parameter features and pixel parameter features with two different MLPs.}
\label{fig:PFG}
\end{figure}

The above four parameters, $p_d$, $p_o$, $p_a$, $p_c$, form a voxel-pixel pair parameter vector $p\in \mathbf{R}^4$ which consists of intrinsic parameters, describing a voxel-pixel pair in regard to their relation. As described in Fig. \ref{fig:PFG}, the four-dimensional voxel-pixel pair parameters will first be fed into a multilayer perceptron (MLP), denoted as ${\rm M}(\cdot)$. In our experimental setting, the dimension is increased from 4 to 16. The operation is expressed as:
\begin{equation}\label{EqPFGPair}
p'={\rm M}(p),
\end{equation}
where ${\rm M}:\mathbf{R}^4\rightarrow\mathbf{R}^{16}$, $p'$ is voxel-pixel pair parameter feature vector. Incidentally, voxel-pixel pair parameter features would be reused in the VPF module.

The next step is to feed the voxel-pixel pair parameter features into another two MLPs, ${\rm M}_v(\cdot)$ and ${\rm M}_p(\cdot)$, generating voxel parameter features and pixel parameter features, respectively, as shown below:
\begin{equation}\label{eq:PFG}
p_v={\rm M}_v(p'), p_p={\rm M}_p(p'),
\end{equation}
where ${\rm M}_v:\mathbf{R}^{16}\rightarrow\mathbf{R}^{C_v}$, $M_p:\mathbf{R}^{16}\rightarrow\mathbf{R}^{C_p}$, $C_v$ denotes the number of channels for a voxel, $C_p$ stands for the number of channels for a pixel, $p_v\in\mathbf{R}^{C_v}$ is voxel parameter features, and $p_p\in\mathbf{R}^{C_p}$ represents pixel parameter features.

\subsubsection{Parameter-Based Weighting Module.}
\begin{figure}[h]
\centering
\includegraphics[width=\columnwidth]{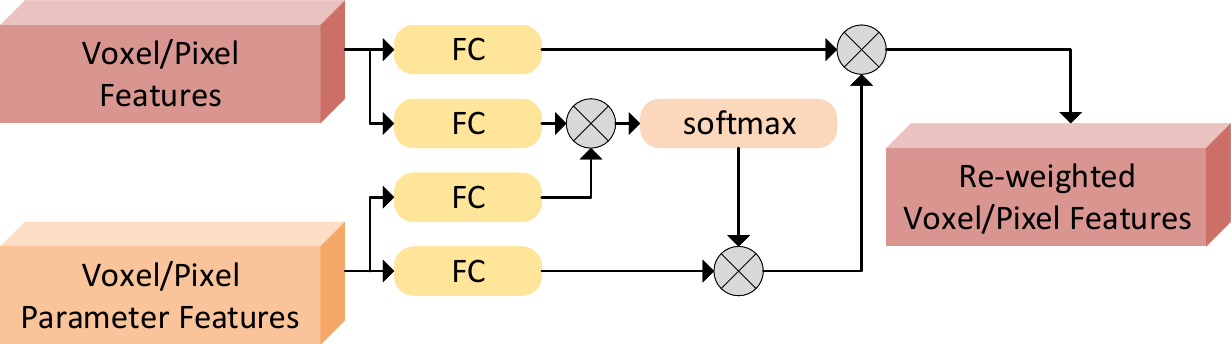} % Reduce the figure size so that it is slightly narrower than the column. Don't use precise values for figure width.This setup will avoid overfull boxes.
\caption{The structure of Parameter-Based Weighting module. The voxel features or pixel features are re-weighted through voxel parameter features or pixel parameter features, respectively.}
\label{fig:PBW}
\end{figure}

After combining the voxel-pixel parameters and generating the parameter features from PFG, these features would be used to initiatory re-weight the voxel features from the LiDAR feature map and the pixel features from the camera feature map, further enhancing the features, as illustrated in Fig. \ref{fig:PBW}.

Features of a voxel coming from the LiDAR feature map are expressed by $b_v\in\mathbf{R}^{C_v}$, whereas features of a pixel coming from the camera feature map are symbolized by $b_p\in\mathbf{R}^{C_p}$. The corresponding voxel parameter feature vector coming from the PFG is represented as $p_v\in\mathbf{R}^{C_v}$, whose dimension is designed the same as $b_v$ intentionally. The re-weighting procedure of voxel features and voxel parameter features is shown as:
\begin{equation}\label{eq:PBWVoxel}
b_v'={\rm softmax}({\rm B}_v^1(b_v)\circ {\rm P}_v^1(p_v))\circ {\rm P}_v^2(p_v )\circ {\rm B}_v^2(b_v),
\end{equation}
where ${\rm B}_v^1(\cdot)$, ${\rm B}_v^2(\cdot)$ and ${\rm P}_v^1(\cdot)$, ${\rm P}_v^2(\cdot)$ are neural networks that preserve the dimensions of $b_v$ and $p_v$, respectively, and $b_v'\in\mathbf{R}^{C_v}$ is the re-weighted voxel features.

In addition, the pixel parameter feature vector coming from the PFG is written as $p_p\in\mathbf{R}^{C_p}$, whose dimension is designed the same as the $b_p$ on purpose. And the re-weighting procedure of pixel features and pixel parameter features is shown as:
\begin{equation}\label{eq:PBWPixel}
b_p'={\rm softmax}({\rm B}_p^1(b_p)\circ {\rm P}_p^1(p_p))\circ {\rm P}_p^2(p_p )\circ {\rm B}_p^2(b_p),
\end{equation}
where ${\rm B}_p^1(\cdot)$, ${\rm B}_p^2(\cdot)$ and ${\rm P}_p^1(\cdot)$, ${\rm P}_p^2(\cdot)$ are neural networks that preserve the dimensions of $b_p$ and $p_p$, respectively, and $b_p'\in\mathbf{R}^{C_p}$ is the re-weighted pixel features.

\subsubsection{Voxel-Pixel Fusion Module.}
\begin{figure}[t]
\centering
\includegraphics[width=\columnwidth]{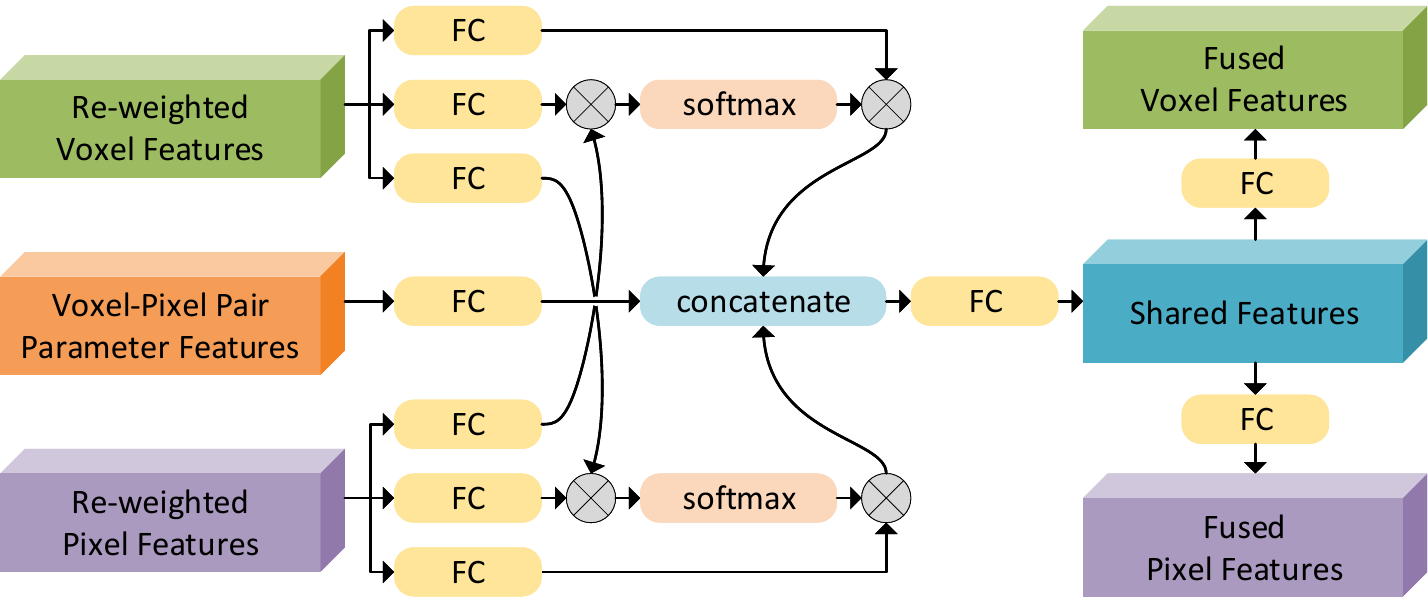} % Reduce the figure size so that it is slightly narrower than the column. Don't use precise values for figure width.This setup will avoid overfull boxes.
\caption{The structure of the Voxel-Pixel Fusion module. The re-weighted voxel features, re-weighted pixel features, and voxel-pixel pair parameter features are combined and transformed into the fused voxel features and pixel features.}
\label{fig:VPF}
\end{figure}
After re-weighting the voxel features and pixel features from PBW, we will fuse cross-sensor features formally in this module. The inputs of the VPF module are the re-weighted voxel features, re-weighted pixel features, and voxel-pixel pair parameter features.

A cross attention mechanism is adopted between the re-weighted voxel features and the re-weighted pixel features to enhance fusion. First, we calculate the attention scores for re-weighted voxel features. As shown in Eq. \ref{eq:VPFTempVoxel}, since the dimensions of $b_v'$ and $b_p'$ can be different, ${\rm B}_p^3(\cdot)$ is a neural network, which makes $b_p'$’s dimension the same as that of $b_v'$. ${\rm B}_v^3(\cdot)$ is also a neural network, which keeps the same dimension of $b_v'$. After Hadamard product of ${\rm B}_v^3(b_v')$ and ${\rm B}_p^3(b_p')$, a softmax is applied to calculate the attention scores. After calculating the attention scores, which are used to adjust the re-weighted voxel features, we further propose ${\rm B}_v^5(\cdot)$ as a neural network which is used with $b_v'$. Next, it is multiplied with the corresponding attention scores, yielding temporary voxel features $b_v^t\in \mathbf{R}^{C_v}$.
\begin{equation}\label{eq:VPFTempVoxel}
b_v^t={\rm softmax}({\rm B}_v^3(b_v')\circ {\rm B}_p^3(b_p'))\circ {\rm B}_v^5(b_v').
\end{equation}

The re-weighted pixel features, whose method of calculating attention scores is very similar to the one shown in Eq. \ref{eq:VPFTempVoxel}. Specifically, ${\rm B}_v^4(\cdot)$ is a neural network, converting $b_v'$’s dimension to one which is the same as that of $b_p'$, and ${\rm B}_p^4(\cdot)$ is also a neural network whose output has the exact dimension of $b_p'$. After calculating the attention scores, which is used to adjust re-weighted pixel features, we further propose ${\rm B}_p^5(\cdot)$ as a neural network that is used with $b_p'$. Next, it is multiplied with the corresponding attention scores, yielding temporary pixel features $b_p^t\in \mathbf{R}^{C_p}$.
\begin{equation}\label{eq:VPFTempPixel}
b_p^t={\rm softmax}({\rm B}_v^4(b_v')\circ {\rm B}_p^4(b_p'))\circ B_p^5(b_p').
\end{equation}

Concerning voxel-pixel pair parameter features $p'$, which comes from PFG, we just use a simple neural network, ${\rm P}(\cdot)$, to generate temporary voxel-pixel pair parameter features, $p^t\in \mathbf{R}^{16}$, which is defined as:
\begin{equation}\label{eq:VPFTempParam}
p^t={\rm P}(p').
\end{equation}

Up to now, the temporary features $b_v^t$, $b_p^t$, and $p^t$ are concatenated, fed into a neural network ${\rm S}(\cdot)$ to generate the shared features $s$, which is denoted as:
\begin{equation}\label{EqVPFShare}
s={\rm S}({\rm CONCAT}(b_v^t,b_p^t,p^t)),
\end{equation}
where ${\rm S}(\cdot)$ keeps the same dimension of the input and $s\in \mathbf{R}^{C_v+C_p+16}$.

Then, the shared features is fed into ${\rm S}_v(\cdot)$, a neural network, to generate the fused voxel features $b_v''$, and fed into ${\rm S}_p(\cdot)$ to generate the fused pixel features, both as shown below:
\begin{equation}\label{EqVPF}
b_v''={\rm S}_v(s), b_p''={\rm S}_p(s),
\end{equation}
where $b_v''\in \mathbf{R}^{C_v}$ has the same dimension of the original voxel features, and $b_p''\in \mathbf{R}^{C_p}$ keeps the same dimension of the original pixel features.

The overall structure of the VPF module is shown in Fig. \ref{fig:VPF}. The fused voxel features and fused pixel features, which serve as residuals, are appended to the original features. Finally, fused feature maps substitute the original feature maps and return to backbone networks.

\subsection{LiDAR Stream}
Voxel-wise feature extraction gets a balance between the one of using raw points or BEV representation since it can preserve 3D space information and adopt convolutional operation. However, because the VPF layer fuses voxel's and pixel's features, it is required to employ a 3D backbone network based on a voxelized point cloud. Therefore, we choose PV-RCNN \cite{pvrcnn}, whose backbone network extracts voxel-wise features, as the LiDAR stream of VPFNet. Besides, the LiDAR stream can be replaced with other 3D object detection networks if they use 3D convolution to extract voxel-wise features in the same way.

\subsection{Camera Stream}
We refer to Cascade RCNN \cite{cascadercnn} as the camera stream of VPFNet, and we exploit ResNeXt-151 \cite{resnext} as the 2D backbone network.
The 2D detection task in the camera stream is auxiliarily used for training because the final 3D detection relies on the LiDAR stream.

\begin{table*}[t]
\centering

\begin{tabular}{c||ccc|ccc|ccc||c}
\hline
\multirow{2}{*}{\centering Method} & \multicolumn{3}{c|}{Car} & \multicolumn{3}{c|}{Pedestrian} & \multicolumn{3}{c||}{Cyclist} & \multirow{2}{*}{\centering mAP} \\
 & Easy & Mod. & Hard & Easy & Mod. & Hard & Easy & Mod. & Hard &  \\ \hline\hline
\textbf{LiDAR-based}: & & & & & & & & & & \\
TANet \cite{tanet} & 84.39 & 75.94 & 68.82 & 53.72 & 44.34 & 40.49 & 75.70 & 59.44 & 52.53 & 61.71 \\
MMLab PV-RCNN \cite{pvrcnn} & \textbf{90.25} & \textbf{81.43} & \textbf{76.82} & 52.17 & 43.29 & 40.29 & 78.60 & 63.71 & 57.65 & 64.91 \\
3DSSD \cite{3dssd} & 88.36 & 79.57 & 74.55 & 54.64 & 44.27 & 40.23 & 82.48 & 64.10 & 56.90 & 65.01 \\
HotSpotNet \cite{hotspotnet} & 87.60 & 78.31 & 73.34 & 53.10 & 45.37 & 41.47 & \textbf{82.59} & \textbf{65.95} & \textbf{59.00} & 65.19 \\ \hline
\textbf{Fusion-based}: & & & & & & & & & & \\
MV3D \cite{mv3d} & 74.97 & 63.63 & 54.00 & - & - & - & - & - & - & - \\
UberATG-MMF \cite{mmf} & 88.40 & 77.43 & 70.22 & - & - & - & - & - & - & - \\
PI-RCNN \cite{pircnn} & 84.37 & 74.82 & 70.03 & - & - & - & - & - & - & - \\
MLOD \cite{mlod} & 77.24 & 67.76 & 62.05 & 47.58 & 37.47 & 35.07 & 68.81 & 49.43 & 42.84 & 54.25 \\
AVOD-FPN \cite{avod} & 83.07 & 71.76 & 65.73 & 50.46 & 42.27 & 39.04 & 63.76 & 50.55 & 44.93 & 56.84 \\
F-PointNet \cite{fpointnet} & 82.19 & 69.79 & 60.59 & 50.53 & 42.15 & 38.08 & 72.27 & 56.12 & 49.01 & 57.86 \\
PointPainting \cite{pointpainting} & 82.11 & 71.70 & 67.08 & 50.32 & 40.97 & 37.87 & 77.63 & 63.78 & 55.89 & 60.82 \\
Faraway-Frustum \shortcite{farawayfrustum} & 87.45 & 79.05 & 76.14 & 46.33 & 38.58 & 35.71 & 77.36	& 62.00 & 55.40 & 62.00 \\
F-ConvNet \cite{fconvnet} & 87.36 & 76.39 & 66.69 & 52.16 & 43.38 & 38.80 & 81.98 & 65.07 & 56.54 & 63.15 \\ \hline\hline
VPFNet (Ours) & 88.51 & 80.97 & 76.74 & \textbf{54.65} & \textbf{48.36} & \textbf{44.98} & 77.64 & 64.10 & 58.00 & \textbf{65.99} \\ \hline
\end{tabular}
%}
\caption{Published comparison on KITTI 3D testing set from KITTI official leaderboard. mAP is the mean of all the categories under all the difficulties.}
\label{TabKITTITest}
\end{table*}

\subsection{Loss Function}
There are two streams in the VPFNet, the LiDAR stream, and the camera stream. We follow the total loss in PV-RCNN \cite{pvrcnn} as the LiDAR stream loss, and the one in Cascade R-CNN \cite{cascadercnn} as the camera stream loss. Therefore, the total loss of VPFNet is integrated of the losses in both streams, which is expressed as:
\begin{equation}\label{EqTotalLoss}
L=L_{LiDAR}+L_{Camera},
\end{equation}
where $L_{LiDAR}$ is the loss of the LiDAR stream, and $L_{Camera}$ is the loss of the camera stream.

\section{Experiment}
\subsection{Experimental Setup}

\subsubsection{Experimental Dataset}
We evaluate our VPFNet on KITTI Vision Benchmark \cite{kitti}, which provides 7,481 training samples and 7,518 testing samples for the 3D and BEV object detection tasks. Each sample contains a LiDAR point cloud, camera RGB image, and calibration matrix. For training samples, labels are a plus. Obstacle categories for evaluation are car, pedestrian, and cyclist. The difficulties are easy, moderate, and hard, representing fully visible and slightly truncated, partly occluded and moderate truncation, challenging to see and severe truncation, respectively. The difference between 3D and BEV object detection tasks is that BEV does not consider the object's height.

\subsubsection{Evaluation Metrics}
KITTI adopts average precision (AP) for evaluating each category under each difficulty, and it is calculated with 40 recall positions. For multi-class evaluation under multi-level difficulty, we further utilize mean average precision (mAP) as the evaluation metric, the mean of APs of all the categories under all the difficulties.

\subsubsection{Training Detail}
Data augmentation can synthesize data artificially by modifying the original one, and this strategy is quite effective and mature for a single sensor. However, it is challenging for fusion-based methods. Therefore, we pre-train the LiDAR stream and camera stream separately with data augmentation enabled. We first pre-train PV-RCNN \cite{pvrcnn} about 60 epochs for the LiDAR stream on the KITTI dataset. Then, we pre-train Cascade R-CNN \cite{cascadercnn} about six epochs for the camera stream. In detail, the learnable weights in PV-RCNN are randomly initialized. On the contrary, Cascade R-CNN uses the pre-trained model of the COCO \cite{coco} dataset and is finetuned on the KITTI dataset. After pre-training, the voxel-pixel fusion layer is joined, and the complete VPFNet continues to be trained about five epochs without using data augmentation. Besides, the total batch size is 24 on eight NVIDIA Tesla V100 GPUs for all the training settings, and the cosine annealing strategy is adopted to adjust the learning rate dynamically. In the pre-training stage, learning rates for both the LiDAR stream and camera stream are set at 0.01. Finally, the VPFNet is trained with a learning rate of 0.002.

\subsection{Result on KITTI Testing Set}
Our VPFNet is trained and evaluated on the KITTI \cite{kitti} dataset. We compare our VPFNet with other state-of-the-art approaches on the testing set on the public leaderboard for fairness. As shown in Table \ref{TabKITTITest}, we compare our VPFNet with both LiDAR-based and fusion-based methods and show that our VPFNet has outperformed all the state-of-the-art methods in mAP for multi-class 3D object detection under multi-level difficulty.

\begin{figure*}[t]
\centering
\includegraphics[width=\textwidth]{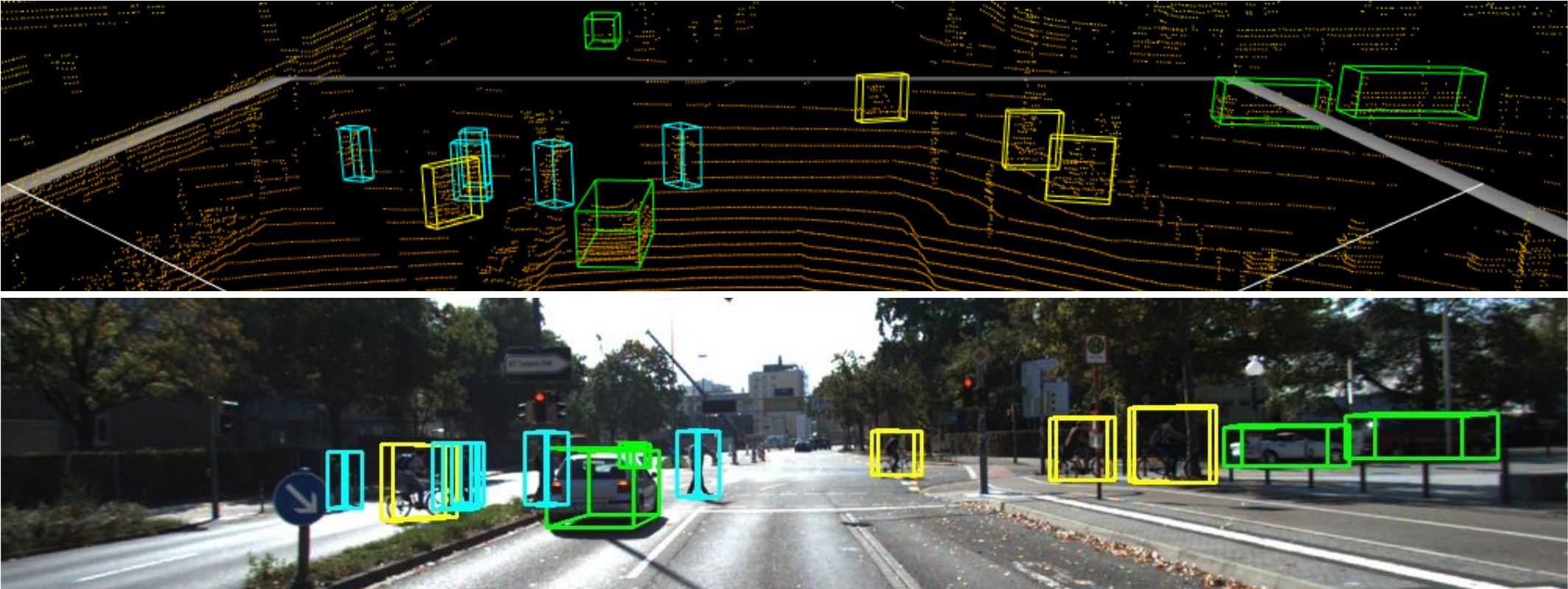} % Reduce the figure size so that it is slightly narrower than the column.
\caption{Visualization of detection of our VPFNet on KITTI validation set. The first row depicts the 3D detection on a LiDAR point cloud. The second one demonstrates the projected 3D detection on the camera RGB image. Cars, pedestrians, and cyclists are detected simultaneously, presented as green, blue, and yellow boxes, respectively.}
\label{fig:result}
\end{figure*}

\subsection{Ablation Study}
Since the KITTI \cite{kitti} dataset does not provide a validation set and only provides a training set and testing set, we split the training set into training split and testing split. Therefore, the original training set, including 7,481 frames, is separated into a 3,712 training split and 3,769 validation split. Furthermore, because KITTI Vision Benchmark has made a policy of submission which requires only one upload to the tuned algorithms, we follow the rule to validate the effect of our proposed method on validation split.

\subsubsection{Effect of Modules within VPF Layer}
\begin{table}[t]
\centering
%\resizebox{.95\columnwidth}{!}{
\begin{tabular}{ccc|ccc|c}
\hline
PFG & \multicolumn{1}{l}{PBW} & \multicolumn{1}{l|}{VPF} & Car & Ped. & Cyc. & mAP \\ \hline
 &  &  & 85.60 & 61.15 & 72.95 & 73.23 \\
\checkmark &  & \checkmark & 85.51 & 63.08 & 76.62 & 75.07\\
\checkmark & \checkmark & \checkmark & \textbf{86.08} & \textbf{64.48} & \textbf{77.06} & \textbf{75.87} \\ \hline
\end{tabular}
\caption{Effect of modules within VPF layer on KITTI 3D validation split. The value for each class is the mAP under easy, moderate, and hard difficulties. The rightest column shows the mAP for all the classes.}
\label{tab:modules}
\end{table}
To verify the effect of each module in the VPF layer, we gradually insert each module and see the outcome performance, as illustrated in Table \ref{tab:modules}. The first experimental item shows the baseline, which does not fuse features between two streams. The second one uses the VPF module, and the PFG module is also required to generate the voxel-pixel pair parameter features. The final one is the complete VPF layer, including PFG, PBW, and VPF modules.
We can notice that when both PFG and VPF modules are enabled, the performance gain is elevated. Moreover, the final performance is the best when all the modules are fully integrated.

\subsubsection{Effect of Arrangement of VPF Layers}
\begin{table}[t]
\centering
%\resizebox{.95\columnwidth}{!}{
\begin{tabular}{ccc|ccc|c}
\hline
1 & \multicolumn{1}{l}{2} & \multicolumn{1}{l|}{3} & Car & Pedestrian & Cyclist & mAP\\ \hline
 &  &  & 85.60 & 61.15 & 72.95 & 73.23\\
\checkmark &  &  & 86.08 & 64.48 & 77.06 & 75.87\\
 & \checkmark &  & \textbf{86.74} & \textbf{65.83} & \textbf{77.45} & \textbf{76.68}\\
 & & \checkmark & 85.93 & 63.34 & 74.38 & 74.55\\
\checkmark & \checkmark &  & 86.24 & 61.54 & 75.25 & 74.34\\ \hline
\end{tabular}
\caption{Effect of arrangement of VPF layers on KITTI 3D validation split. The value for each class is the mAP under easy, moderate, and hard difficulties. The rightest column shows the mAP for all the classes.}
\label{tab:arrange}
\end{table}
Since VPF layers can be a bridge connecting each down-sampling layer or repeated inserted, we conduct experiments to discover the best choice. In the beginning, we insert the VPF layer to the first, the second, or the third down-sampling layer. The results show that bridging the VPF layer on the second down-sampling layer pair for all the classes is the best choice. Finally, we try to insert VPF layers to both the first and second down-sampling layers. However, we do not observe any favorable.

\subsection{Visualization}
In order to prove the effectiveness of our proposed modules, we visualize the 3D detection of our method on a LiDAR point cloud and the corresponding image. As shown in Fig. \ref{fig:result}, our proposed VPFNet, a fusion-based approach, can detect multi-class objects simultaneously in a complicated environment.

\section{Conclusion}
This paper proposes a novel VPFNet to detect 3D objects, especially for detecting multi-class objects simultaneously and more challenging categories like pedestrians. VPF layer is the core of VPFNet, which fuses features according to the geometric relation and several parameters of a voxel-pixel pair. Extensive experiments validate the effectiveness of our proposed method, which has outperformed the state-of-the-art methods for the multi-class 3D object detection task under multi-level difficulty.

\appendix
% Use \bibliography{yourbibfile} instead or the References section will not appear in your paper
\bibliography{aaai22}
\end{document}